\pdfoutput=1

\documentclass[letterpaper, 10 pt, conference]{ieeeconf} 
\usepackage{times} 

\usepackage{xspace} 
\makeatletter
\DeclareRobustCommand\onedot{\futurelet\@let@token\@onedot} 
\def\@onedot{\ifx\@let@token.\else.\null\fi\xspace} 
 
\let\NAT@parse\undefined
\makeatother

\makeatletter
\newcommand\footnoteref[1]{\protected@xdef\@thefnmark{\ref{#1}}\@footnotemark}
\makeatother

\usepackage[numbers,sort&compress]{natbib}
\usepackage{algorithm}
\usepackage[algo2e,lined]{algorithm2e}
\usepackage{multicol}
\usepackage{enumerate}
\usepackage{color}
\usepackage{colortbl}
\usepackage{graphicx} 
\usepackage{amsmath} 
\usepackage{amssymb}  
\usepackage{xcolor}
\usepackage{mdframed}
\usepackage{subcaption}
\usepackage[utf8]{inputenc}
\usepackage{siunitx}
\usepackage{multirow}
\usepackage[font={small}]{caption}
\usepackage{standalone}
\usepackage{epstopdf}
\usepackage{verbatim}
\usepackage{booktabs}
\usepackage{bm}
\usepackage{flushend}
\usepackage{listings}
\usepackage{makecell}
\usepackage{capt-of}

\SetKwInOut{Parameter}{Parameters}

\newcounter{parms}

\newcommand{\reffig}[1]{Fig.~\ref{#1}}
\newcommand{\refeq}[1]{Eq.~(\ref{#1})}
\newcommand{\reftab}[1]{Table~\ref{#1}}
\newcommand{\refsec}[1]{Section~\ref{#1}}

\definecolor{todo-red}{RGB}{200,12,12}
\definecolor{green4}{RGB}{0,128,0}

\newcommand{\NOTE}[1]{\textcolor{green4}{\textbf{NOTE:} [{#1}]}}

\title{\LARGE \bf
maplab: An Open Framework for Research\\ in Visual-inertial Mapping and Localization
}
\author{\authorblockN{
Thomas Schneider\authorrefmark{1}\authorrefmark{2}, 
Marcin Dymczyk\authorrefmark{1}\authorrefmark{2}, 
Marius Fehr\authorrefmark{1}\authorrefmark{2},  \\
Kevin Egger\authorrefmark{2}, 
Simon Lynen\authorrefmark{2}\authorrefmark{4},
Igor Gilitschenski\authorrefmark{3}, 
Roland Siegwart\authorrefmark{2}}
\authorblockA{\authorrefmark{2}Autonomous Systems Lab, ETH Z\"urich, 
\authorrefmark{3}CSAIL, MIT, 
\authorrefmark{4}Google Inc., Z\"urich \\
\authorrefmark{1}contributed equally}
}

\begin{document}

\maketitle

%

\begin{abstract}
Robust and accurate visual-inertial estimation is crucial to many of today's challenges in robotics.
Being able to localize against a prior map and obtain accurate and drift-free pose estimates can push the applicability of such systems even further.
Most of the currently available solutions, however, either focus on a single session use-case, lack localization capabilities or an end-to-end pipeline.
%
%
We believe that only a complete system, combining state-of-the-art algorithms, scalable multi-session mapping tools, and a flexible user interface, can become an efficient research platform.

We therefore present maplab, an open, research-oriented visual-inertial mapping framework for processing and manipulating multi-session maps, written in C++.
On the one hand, maplab can be seen as a ready-to-use visual-inertial mapping and localization system.
On the other hand, maplab provides the research community with a collection of multi-session mapping tools that include map merging, visual-inertial batch optimization, and loop closure. 
Furthermore, it includes an online frontend that can create visual-inertial maps and also track a global drift-free pose within a localization map.
In this paper, we present the system architecture, five use-cases, and evaluations of the system on public datasets.
The source code of maplab is freely available for the benefit of the robotics research community.
\end{abstract}

\section{Introduction}

%
%
The ever growing deployment of simultaneous localization and mapping (SLAM) systems poses novel challenges for the robotics community. 
Availability of precise, drift-free pose estimates both outdoors and indoors has become a vital requirement of numerous robotics applications, such as navigation or manipulation.
The increasing popularity of visual-inertial estimation systems created a strong incentive to improve their robustness to viewpoint and appearance changes (daylight, weather, seasons, etc.) or rapid motion.
Current research efforts aim to collect data using heterogeneous agents, build maps of larger scale, cover various visual appearance conditions and maintain maps over a long time horizon.
Investigating these and many related challenges requires a multi-session end-to-end mapping system that can be easily deployed on various robotic platforms and provides ready-to-use algorithms with state-of-the-art performance.
At the same time it needs to offer high flexibility necessary for conducting research.

%
%
Most openly available frameworks for visual and visual-inertial SLAM either focus on a single-session case~\cite{leutenegger2015keyframe} or only provide large-scale batch optimization without an online frontend~\cite{snavely2006photo}.
Usually, they are crafted for a very specific pipeline without a separation between the map structure and algorithms.
They often lack completeness and will not offer a full workflow such that a map can be created, manipulated, merged with previous sessions and reused in the frontend within a single framework.
This impairs the flexibility of such systems, a key for rapid development and research.

\begin{figure}
\centering
\includegraphics[width=1.0\linewidth]{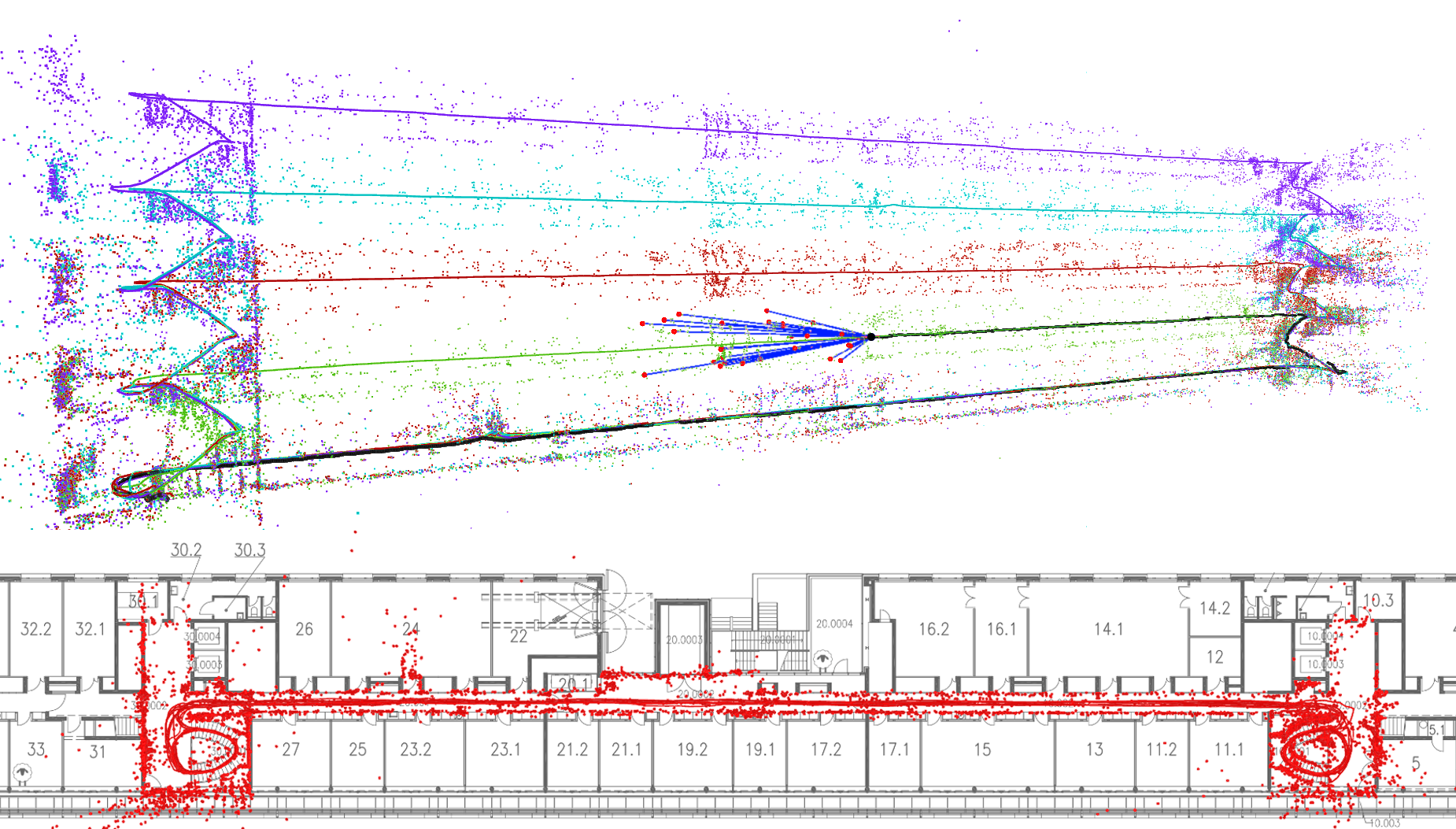}
\caption{
    The maplab framework can build consistent visual-inertial maps from multiple mapping sessions.
    Here, 4~separate sessions are merged and jointly refined.
    The global map can then be used by odometry and localization frontend to correct for any drift when revisiting the area.
    The floorplan is overlayed with the landmarks of all floors demonstrating the accuracy and consistency of the map alignment.
}
\label{fig:teaser}
\end{figure}

%
%
This work addresses this problem by introducing maplab\footnotemark[1], an open visual-inertial mapping framework, written in C++.
\footnotetext[1]{Maplab is available at: \url{www.github.com/ethz-asl/maplab}}
In contrast to existing visual-inertial SLAM systems, maplab does not only provide tools to create and localize from visual-inertial maps but also provides map maintenance and processing capabilities.
These capabilities are offered as a set of tools accessible in a convenient console that can easily be extended through a plugin system.
These tools involve multi-session merging, sparsification, loop closing, dense reconstruction and visualization of maps.
Additionally, maplab includes ROVIOLI (ROVIO with Localization Integration), a mapping and localization frontend based on ROVIO~\cite{bloesch2015robust}, a patch-based visual-inertial odometry system.

%
%
Maplab has been extensively field tested and has been deployed on a variety of robotic platforms including micro aerial vehicles~\cite{burri2015real}, autonomous planes~\cite{hinzmann2016monocular,hinzmann2016robust}, autonomous cars~\cite{buerkiIROS2016}, autonomous underwater vehicles~\cite{traffelet2016target}, and walking robots~\cite{fankhauser2016collaborative}.
It has also served as a research platform for map summarization~\cite{dymczyk2015gist,dymczyk2015keep,dymczyk2016map,dymczyk2016erasing}, map quality evaluation~\cite{merzic2017map}, multi-session 3d reconstruction~\cite{fehr2016reshaping}, topological mapping~\cite{blochliger2017topomap}, visual localization~\cite{gehrig2017visual,oleynikova2015real,lynen2014placeless}, and decentralized mapping~\cite{cieslewski2015map}.

%
%
To the best of our knowledge, maplab is the first visual-inertial mapping framework that integrates a wide variety of use-cases within a single system.
Maplab is free, open-source, and has already proved to be of great use for various research and industry projects.
We strongly believe that the robotics community will harness it both as an off-the-shelf mapping and localization solution, as well as a mapping research testbed.
The contributions of this work can be summarized as follows:
\begin{itemize}
    \item it introduces a general purpose visual-inertial mapping framework using feature-based maps with multi-session support;
    \item it introduces ROVIOLI, a robust visual-inertial estimator tightly coupled with a localization system;
    \item it presents examples of algorithms and data structures for modifying and maintaining maps including map merging, sparsification, place recognition, and visualization;
    \item it highlights the extensibility of the system that makes it well suited for research;
    \item it provides evaluation of selected components of the framework.
\end{itemize}

\section{Related work}
%
%
There are several openly available visual and visual-inertial SLAM systems.
One of the earliest examples is PTAM~\cite{klein2007parallel}, a lightweight approach for mapping and tracking a local map in parallel.
It was originally developed for augmented reality applications so it offers neither large-scale localization nor any offline processing tools.
More recent examples include OKVIS~\cite{leutenegger2015keyframe}, a visual-inertial keyframe-based estimator.
This approach tracks a local map built from recently acquired keyframes, which minimizes the drift locally.
Similarly, semi-dense~\cite{forster2014svo} and dense~\cite{engel2014lsd} odometry frameworks achieve high-quality pose estimates by using photometric error formulations instead of feature-based matching.
None of these methods, however, supports global localization against a previously recorded map. 

%
%
ORB-SLAM~\cite{mur2015orb} and ORB-SLAM2~\cite{MurArtal2017} are vision-based frameworks that offer the possibility to create a map of the environment and then reuse it in a consecutive session, which closely relates to the workflow we propose here.
In contrast to these systems, maplab offers an offline processing toolkit centered around a console user interface, which guarantees high flexibility and permits users to add their own extensions or modify the processing pipelines.
We consider the ability to merge multiple mapping sessions into a single, consistent map and to refine it using a visual-inertial least-squares optimization a core capability of maplab that differentiates it from ORB-SLAM.
Another difference worth emphasizing is the online frontend of maplab, ROVIOLI.
Using image intensity within patches instead of point features guarantees a high level of robustness, even in the presence of motion blur~\cite{bloesch2015robust}.

%
%
Incorporating the capability to process multiple maps has received considerable attention in the SLAM research community with~\cite{Bosse2004} being one of the earliest works incorporating multiple maps in a hybrid metric-topological approach to multi-session mapping.
Use of anchor nodes to stitch together posegraphs from multiple mapping sessions is proposed in~\cite{McDonald2013}.
Trying to establish topological associations between maps is also proposed in~\cite{Churchill2013}, where maps are stored as a set of experiences. 
In contrast, maplab stores a unified localization map allowing to use a carefully selected subset of features, e.g. based on the current appearance conditions~\cite{buerkiIROS2016}.

%
%
Systems that aim to reconstruct the 3d structure from large collections of unordered images~\cite{snavely2006photo,theia-manual,moulon2016openmvg} also contain functionalities similar to maplab.
They typically offer efficient implementations of large-scale bundle adjustment optimization and advanced image and feature matching techniques.
They lack, however, algorithms that process inertial data and cannot be run directly on a robotic platform in order to provide pose estimates online.

\section{The maplab framework}
\label{sec:maplab_overview}
%
%
From the user perspective, the framework consists of two major parts:
\begin{enumerate}[i.]
    \item The online \textbf{VIO and localization frontend}, ROVIOLI, that takes raw visual-inertial sensor data. 
    It outputs (global) pose estimates and can be used to build visual-inertial maps. 
    \item The (offline) \textbf{maplab-console} that lets the user apply various algorithms on maps in an offline batch fashion.
    It does also serve as a research testbed for new algorithms that operate on visual-inertial data.
\end{enumerate}

The maplab framework follows an extensible and modular design.
All software components are organized in packages, which are built using catkin, the official build system of ROS~\cite{quigley2009ros}.
The C++11 standard is used throughout the framework and third-party dependencies are limited to popular and well-maintained libraries, among others Eigen~\cite{eigenweb} for linear algebra and Ceres~\cite{ceres_solver} for non-linear optimization.
Additionally, the framework provides ROS interfaces to conveniently input raw sensor data and output the results, such as pose estimates for an easy deployment on a robotic systems.
The framework uses RViz as a 3d visualization tool to both visualize the state of the online mapping algorithms and the results of the offline processing from the maplab console.

\subsection{Notation}
Throughout this document and the source-code, we use the notation as defined in this section.
A transformation matrix $\mathbf{T}_{AB}\in\text{SE}(3)$ takes a vector $_{B}{\mathbf{p}}\in\mathbb{R}^3$ from the frame of reference $\mathcal{F}_B$ to the frame of reference $\mathcal{F}_A$. 
It can be partitioned into a rotation matrix $\mathbf{R}_{AB}\in\text{SO}(3)$ and a translation vector $_{A}{\mathbf{p}}_{AB}\in\mathbb{R}^3$ as:
\begin{equation}
\label{eq:trafo}
\begin{bmatrix}
_{A}{\mathbf{p}} \\ 1
\end{bmatrix}
 = \mathbf{T}_{AB} \cdot 
\begin{bmatrix}
_{B}{\mathbf{p}} \\ 1
\end{bmatrix}
= 
\begin{bmatrix}
\mathbf{R}_{AB} & _{A}{\mathbf{p}}_{AB}\\ 
 \mathbf{0} & 1 
\end{bmatrix}
\cdot 
\begin{bmatrix}
_{B}{\mathbf{p}} \\ 1
\end{bmatrix}
\end{equation}
The operator $\mathbf{T}_{AB}(\cdot)$ is defined to transform a vector in $\mathbb{R}^3$ from $\mathcal{F}_B$ to the frame of reference $\mathcal{F}_A$ as $_{A}{\mathbf{p}} = \mathbf{T}_{AB} \left( _{B}{\mathbf{p}} \right)$ according to \refeq{eq:trafo}.

\subsection{Workflow for multi-session mapping and localization}
\begin{figure}[h]
\centering
\includegraphics[width=0.75\linewidth]{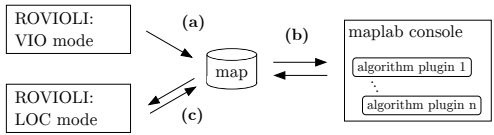}
\caption{Typical workflow in maplab: (a) In VIO~mode, ROVIOLI estimates the pose of an agent w.r.t. a (drifting) local frame; additionally a map is built based on these estimates. (b) Resulting maps can be loaded in the maplab-console where all of the available algorithms can be applied, e.g. map alignment and merging, VI optimization, loop-closure. (c) In LOC~mode, ROVIOLI can load the updated map to track a global (drift-free) pose online.}
\label{fig:maplab_dataflow}
\end{figure}

The typical workflow for a mapping and localization session within the maplab system is illustrated in~\reffig{fig:maplab_dataflow}.
Often, it is beneficial to build a single localization map from multiple mapping sessions to ensure a good spatial and temporal (i.e. different appearances) coverage of the area.
An initial, open loop map is built in each session using ROVIOLI in VIO~mode and stored to disk.
The maps can then be refined using various (offline) tools such as loop closure detection, visual-inertial optimization or co-registration of multiple sessions (map merging).
Detailed inspection of the maps is possible using a large set of different visualizations, statistics and queries.
More advanced modules allow, e.g., to create a dense representation (TSDF, occupancy, etc.) of the environment using data from a depth sensor or from stereo.

The resulting (multi-session) map can then be exported as a compact localization map and used by ROVIOLI (in LOC~mode) for online localization during a second visit to the same place.
Continuous online localization enables accurate tracking of a global pose w.r.t. a known 3d structure and thus compensates for drift in the visual-inertial state estimation.

\subsection{maplab console: the offline user interface}

The maplab framework uses a console user interface to manipulate maps offline.
Multiple maps can be loaded into the console simultaneously, facilitating multi-session mapping experiments.
All algorithms are available through console commands and can be applied to the loaded maps.
Parameters specific to each algorithm are set by console flags or a flag file and can be modified at runtime.
Combined with the real-time visualization of the map in RViz, this greatly facilitates algorithm prototyping and parameter tuning.
It is possible to combine multiple algorithms and experiment with entire processing pipelines.
Changes can be easily reverted by saving and reloading intermediate states of a map from disk.

The console uses a plugin architecture\footnote{\label{fn:wiki}For more details, tutorials and documentation, please visit our wiki page: \url{www.github.com/ethz-asl/maplab/wiki}} and automatically detects all available plugins within the build workspace at run time.
Therefore, the integration of a new algorithm or functionality is possible without any changes to the core packages.
For algorithms that operate on the standard visual-inertial map datatype (see~\refsec{framework:map_structure}), no interfacing work will be necessary.
%

%
%

\subsection{Map structure}
\label{framework:map_structure}

The framework uses a data structure, called VI-map, for visual-inertial mapping data.
The VI-map contains the raw measurements of all sensors and a sparse reconstruction of the covered environment.
Each map may contain multiple \textit{missions} where each is based on a single recording session.
The core structure of a mission is a graph consisting of \textit{vertices} and \textit{edges}.
A \textit{vertex} corresponds to a state captured at a certain point in time.
It contains a state estimate (pose $\mathbf{T}_{MI_k}$, IMU biases, velocity) and visual information from the (multi-)camera system including keypoints, descriptors (BRISK~\cite{leutenegger2011brisk} or FREAK~\cite{alahi2012freak}), tracking information and images.
An \textit{edge} connects two neighboring vertices.
While there are a few different types of edges in maplab, the most common type is the IMU edge.
It contains the inertial measurements recorded between the vertices that the edge connects.
Visual observations tracked by multiple vertices are triangulated as 3d landmarks.
The landmark itself is stored within the vertex that first observed it.
Loop-closures might link observations of one mission to a landmark stored in another mission.

\begin{figure}[]
\centering
\includegraphics[width=1.0\linewidth]{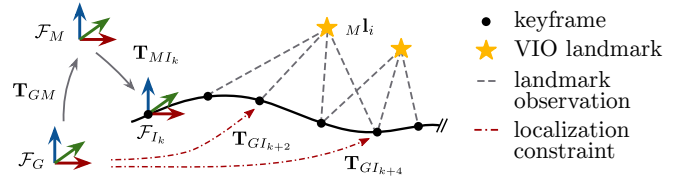}
\caption{
Coordinate frames used in maplab and ROVIOLI:
 $\mathcal{F}_G$: global, gravity-aligned map frame; all missions are anchored in this frame.
 $\mathcal{F}_{M_k}$: gravity-aligned frame that represents the origin of a mission $k$ equivalent to the origin of the VIO.
 $\mathcal{F}_{I_k}$: IMU frame at time stamp $k$ (body frame).
}
\label{fig:frames}
\end{figure}

\reffig{fig:frames} illustrates the map structure and introduces the relevant coordinate frames.
Each mission is anchored in the global coordinate frame $\mathcal{F}_G$ using a transformation $\mathbf{T}_{GM_i}$.
The poses $\mathbf{T}_{M_{i}I_{j}}$ of mission $i$ are expressed w.r.t. the mission frame $\mathcal{F}_{M_i}$.
Therefore, it suffices to manipulate the transformation $\mathbf{T}_{GM_i}$ to anchor multiple missions in a single global coordinate system without the need for updating any vertex poses or landmark positions.

The map structure can be serialized to the Google Protobuf format, enabling portable file serialization and network transmission.
Furthermore, data-intensive objects (such as images, dense reconstructions, etc.) can be attached to the maps using a resource management system.
Resources are linked to either a \textit{vertex} or a set of \textit{missions} or simply a timestamp, and are stored on the file system separate from the main mapping data.
This architecture allows for (cached) loading such (potentially large) objects on demand, effectively reducing the peak memory usage.
This facilitates research in areas such as dense reconstruction and image-based/enhanced localization on large-scale maps that might otherwise exhaust the available memory on certain platforms.


\subsection{Core packages of maplab}
\label{framework:core}

The maplab framework incorporates implementations of several state-of-the-art algorithms.
All of them are conveniently accessible from the maplab console.
We only briefly highlight the ones that, in our opinion, bring a particular value to the robotics community:

\textbf{VIWLS:} visual-inertial weighted least-squares optimization with cost terms similar to~\cite{leutenegger2015keyframe}.
The main batch optimization algorithm of the framework is used to refine maps e.g. after initialization with ROVIOLI or after loop-closures have been established.
By default, the optimization problem is constructed using visual and inertial data, but optionally it can include wheel odometry, GPS measurements or other types of pose priors.

\textbf{Loop closure/localization:} a complete loop closure and localization system based on binary descriptors.
The search backend uses an inverted multi-index for efficient nearest neighbor retrieval on projected binary descriptors. 
The algorithm is a (partial) implementation of \cite{lynen2015get}.

\textbf{ROVIOLI:} online visual-inertial mapping and localization frontend, see~\refsec{framework:rovioli} for details.

\textbf{Posegraph relaxation:} posegraph optimization using edges introduced by the loop closure system. 
The algorithm is similar to~\cite{sunderhauf2012switchable}.
Optionally, a Cauchy loss might be used to increase the robustness against false loop closures.

\textbf{aslam\_cv2:} a collection of computer vision data structures and algorithms.
It includes various camera and distortion models as well as algorithms for feature detection, extraction, tracking and geometric vision.

\textbf{Map sparsification:} algorithms to select the best landmarks for localization~\cite{dymczyk2015gist,dymczyk2015keep} and keyframe selection to sparsify the pose graph.
Useful for processing large-scale maps or for lifelong mapping.

\textbf{Dense reconstruction:} a collection of dense reconstruction, depth fusion and surface reconstruction~\cite{oleynikova2016voxblox} algorithms. Also includes an interface to CMVS/PMVS2~\cite{Furu:2010}. See~\refsec{dense_reconstruction} for details.

\subsection{ROVIOLI: online VIO and localization frontend}
\label{framework:rovioli}
ROVIOLI (ROVIO with Localization Integration) is maplab's mapping and localization frontend which is used to build maps from raw visual and inertial data and also localize w.r.t. existing maps online.
It is built around the visual-inertial odometry framework ROVIO \cite{bloesch2015robust} and extends it with localization and mapping capabilities.
The following two modes of operation are available: (i) \textit{VIO~mode (Visual Inertial Odometry)} in which a map is built based on the VIO estimates and (ii) \textit{LOC~mode} where additionally localization constraints are processed to track a (drift-free) global pose estimate w.r.t. a given map.
The localization maps are either created directly in a previous (single-session) of ROVIOLI or are exported from the maplab-console.
The preparation of a localization map within the console allows for building complex processing pipelines (e.g. multi-session maps, data selection and compression).

\begin{figure}
\centering
\includegraphics[width=1.0\linewidth]{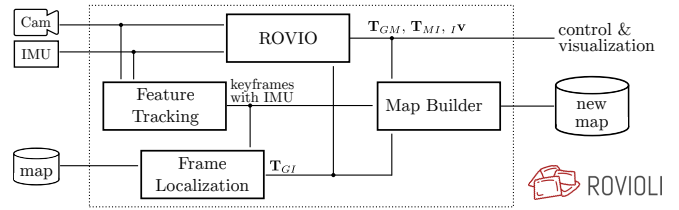}
\caption{Modules and data flows within ROVIOLI (ROVIO~\cite{bloesch2015robust} with Localization Integration).}
\label{fig:rovioli_flow}
\end{figure}

An overview of the (main) data flows and modules within ROVIOLI are shown in~\reffig{fig:rovioli_flow}.
The \textit{Feature Tracking} module detects and tracks BRISK~\cite{leutenegger2011brisk} or FREAK~\cite{alahi2012freak} keypoints.
Feature correspondences between frames are established by matching descriptors from frame to frame.
The expected matching window is predicted based on integrated gyroscope measurements to increase the efficiency and robustness.
In \textit{LOC~mode}, keyframes containing feature points and descriptors are processed by the \textit{Frame Localization} module to establish 2d-3d matches against the provided localization map.
These 2d-3d matches are used to obtain a global pose estimate $\mathbf{T}_{GI_k}$ w.r.t. the map's frame of reference (see~\reffig{fig:frames}) using a P3P algorithm within a RANSAC scheme.
The raw global pose estimates are fed to ROVIO where they are fused with the odometry constraints to estimate a transformation $\mathbf{T}_{GM}$ in addition to the local odometry pose $\mathbf{T}_{MI}$.
The outputs of all modules are synchronized within the \textit{Map Builder} to construct a visual-inertial map (VI-map).
The resulting map can serve as a localization map in subsequent sessions or can be loaded into the maplab console for further processing.

A process-internal publisher-subscriber data exchange layer manages the data flows between all modules within ROVIOLI.
This architecture makes it easy to extend the current online pipeline with new algorithms, e.g. for online multiagent mapping, semantic SLAM, or localization research.


\section{Use-cases}
\label{sec:maplab_usecases}
%
%
%
This section gives an overview of five common use cases of maplab: online mapping and localization, multi-session mapping, map maintenance, large-scale mapping and dense reconstruction.
While maplab offers much more than that, we believe these examples highlight the capabilities of the system, the expected performance and its scalability.

Furthermore, we provide the related console commands to reproduce every example.
The intention is to show that the following results can be obtained by relying solely on the user interface, without any additional code development.
For more documentation, updated commands, datasets and tutorials, please visit our wiki page: \url{www.github.com/ethz-asl/maplab/wiki}\,.

\subsection{Online mapping and localization with ROVIOLI}
%
%
For many robotic applications it is of high importance to have access to (drift-free) global pose estimates.
Such capability enables, e.g., teach\&repeat scenarios, robotic manipulation and precise navigation.
Within maplab, as a first step, we use ROVIOLI to create an initial VI-map of the desired area of operation.
The sensor data can be provided either offline in a Rosbag or online using ROS topics.
Upon completion, the VI-map is automatically loop-closed, optimized and optionally keyframed and summarized to obtain a compact localization map. 
In a second session the localization map can be passed to ROVIOLI to obtain drift-free global pose estimates in the mapped area.

\begin{figure}
    \centering
    \includegraphics[width=0.49\textwidth]{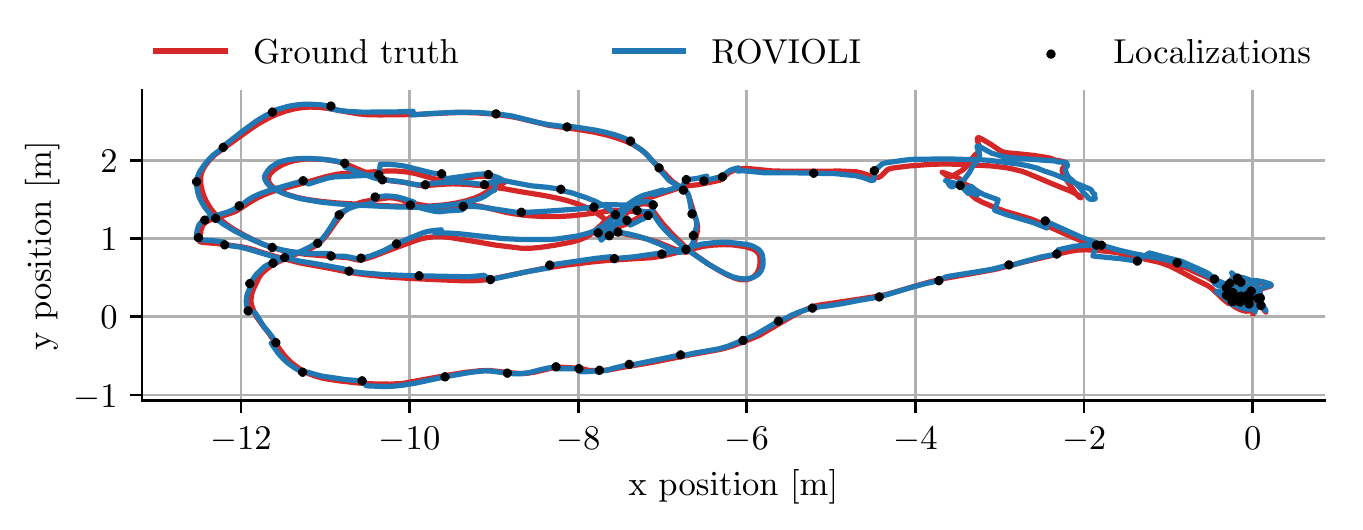}
    \includegraphics[width=0.49\textwidth]{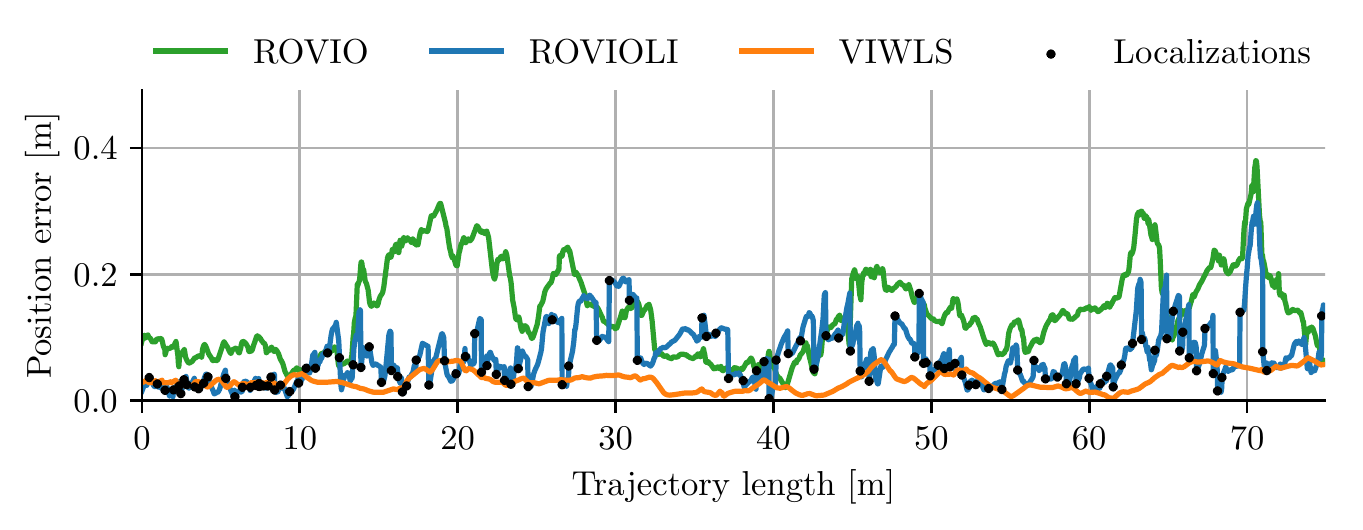}
    \caption{Evaluation of ROVIOLI on the EuRoC machine hall dataset~\cite{burri2016euroc}. \textit{Top}: Ground-truth positions overlayed with the ROVIOLI position estimates. \textit{Bottom}: Position error of the visual-inertial odometry pipeline ROVIO~\cite{bloesch2015robust}, ROVIOLI and the optimized VI-map (VIWLS) compared to the ground truth.}
    \label{fig:rovio_euroc_evaluation_error}
\end{figure}

We evaluated the ROVIOLI estimates against plain ROVIO~\cite{bloesch2015robust} results and the estimates from a full-batch optimization on the EuRoC datasets~\cite{burri2016euroc}.
To that end, in a first step, we created a localization map using one of the datasets.
Then in a second step we processed a second EuRoC dataset using both ROVIOLI (using the previously built map) and ROVIO.
The results are presented in~\reffig{fig:rovio_euroc_evaluation_error} and~\reftab{table:rovio_euroc_evaluation_rmse}, where we compare the groundtruth error of ROVIO, ROVIOLI, and the full-batch optimized trajectory.
These experiments demonstrate the drift-free performance of the system and the improvements upon the regular VIO estimation.
Additionally, \reftab{table:rovioli_timing_full} shows timing information of ROVIO and ROVIOLI compared to ORB-SLAM2~\cite{MurArtal2017}.

%
%

\begin{table}[]
\scriptsize
\centering
\caption{Global position and orientation RMSEs on EuRoC datasets~\cite{burri2016euroc} for ROVIO (only VIO), ROVIOLI using one of the datasets as a localization map and ROVIO+VIWLS that corresponds to a full batch visual-inertial least-squares optimization (VIWLS).
Additionally, the results of ORB-SLAM2~\cite{MurArtal2017} (in batch and real-time) are compared. 
ROVIO and ROVIOLI use a single camera and IMU data whereas ORB-SLAM2 uses a stereo camera.
The localization map for ORB-SLAM2 has been built in SLAM mode whereas the localization evaluation has been performed in localization mode.
For V2-medium, we were unable to build a map with ORB-SLAM2's real-time mode as the estimator diverged (marked with X).}
\label{table:rovio_euroc_evaluation_rmse}
\begin{tabular}{@{}l|rr|rr}
\toprule
 &
\multicolumn{2}{c|}{\begin{tabular}[c]{c}MH1\\ *LOC: MH2\end{tabular}} & 
\multicolumn{2}{c}{\begin{tabular}[c]{c}V2-easy\\ *LOC: V2-medium\end{tabular}} \\
\cline{2-5}
 & 
\multicolumn{1}{c}{position} & 
\multicolumn{1}{c|}{orientation} & 
\multicolumn{1}{c}{position} & 
\multicolumn{1}{c}{orientation} \\
\hline
ROVIO & 
\begin{tabular}[c]{@{}r@{}l@{}} 
    0.178 & \,m\\ 
\end{tabular} & 
\begin{tabular}[c]{r@{}l@{}}
    1.49 & \,deg \\
\end{tabular} & 
\begin{tabular}[c]{@{}r@{}l@{}}
    0.064 & \,m\\ 
\end{tabular} & 
\begin{tabular}[c]{@{}r@{}l@{}}
    0.90 & \,deg \\
\end{tabular} \\
\hline
\begin{tabular}[c]{@{}l@{}}ROVIOLI*\end{tabular} & 
\begin{tabular}[c]{@{}r@{}l@{}}
    0.082 & \,m \\
\end{tabular} & 
\begin{tabular}[c]{@{}r@{}l@{}}
    1.43 & \,deg \\
\end{tabular} & 
\begin{tabular}[c]{@{}r@{}l@{}}
    0.057 & \,m\\ 
\end{tabular} & 
\begin{tabular}[c]{@{}r@{}l@{}}
    1.57 & \,deg \\
\end{tabular} \\
\hline
\begin{tabular}[c]{@{}l@{}}ROVIO+\\ VIWLS\end{tabular} & 
\begin{tabular}[c]{@{}r@{}l@{}}
    0.036 & \,m\\ 
\end{tabular} & 
\begin{tabular}[c]{@{}r@{}l@{}}
    1.29 & \,deg \\
\end{tabular} & 
\begin{tabular}[c]{@{}r@{}l@{}}
    0.027 & \,m\\ 
\end{tabular} & 
\begin{tabular}[c]{@{}r@{}l@{}}
    1.06 & \,deg \\
\end{tabular} \\
\hline
\begin{tabular}[c]{@{}l@{}}ORB-SLAM2*\\ (batch mode)\end{tabular} & 
\begin{tabular}[c]{@{}r@{}l@{}}
    0.084 & \,m\\ 
\end{tabular} & 
\begin{tabular}[c]{@{}r@{}l@{}}
    0.78 & \,deg \\
\end{tabular} & 
\begin{tabular}[c]{@{}r@{}l@{}}
    0.121 & \,m\\ 
\end{tabular} & 
\begin{tabular}[c]{@{}r@{}l@{}}
    1.14 & \,deg \\
\end{tabular} \\
\hline
\begin{tabular}[c]{@{}l@{}}ORB-SLAM2*\\ (real-time)\end{tabular} & 
\begin{tabular}[c]{@{}r@{}l@{}}
    0.464 & \,m\\ 
\end{tabular} & 
\begin{tabular}[c]{@{}r@{}l@{}}
    13.34 & \,deg \\
\end{tabular} & 
    X &
    X \\
\bottomrule
\end{tabular}
\end{table}

\begin{table}[]
\scriptsize
\centering
\caption{\textit{(a)} Timing and CPU load for ROVIO, ROVIOLI and ORB-SLAM2 on EuRoC MH1 dataset processed at 20\,Hz.
In case of ROVIOLI and ORB-SLAM2 (marked with *), the estimator was set to localize against a map built from EuRoC MH2.
All reported values have been measured on an Intel Xeon E3-1505M@2.8Ghz.
A CPU load of 800\% corresponds to fully utilizing \textbf{all} 8 (logical) cores of the CPU.
\textit{(b)} Single frame processing times for the individual blocks of ROVIOLI.
The total time does not correspond to the sum of the individual blocks as they run in parallel.
Instead, it is the time it takes for a single frame to be fully processed.
}
\label{table:rovioli_timing_full}

\begin{subtable}{0.63\linewidth}
\centering
\caption{} 
\begin{tabular}{@{}l|r|c@{}}
\toprule
    & \shortstack{Frame \\ update} & CPU load \\
\hline
    ROVIO &
    23 \,ms&
    \begin{tabular}{@{}r@{}c@{}l@{}}
        56\%&$\pm$&7.7\% \\
    \end{tabular} \\
\hline
    ROVIOLI* &
    44 \,ms& 
    \begin{tabular}{@{}r@{}c@{}l@{}}
        105\%&$\pm$&14.8\% \\
    \end{tabular} \\
\hline
    \begin{tabular}[c]{@{}l@{}}ORB-SLAM2*\\ (batch mode) \end{tabular} & 
    63 \,ms& 
    \begin{tabular}{@{}r@{}c@{}l@{}}
        162\%&$\pm$&10.9\% \\
    \end{tabular} \\
\bottomrule
\end{tabular}
\end{subtable} %
\begin{subtable}{0.35\linewidth}
\centering
\caption{}
\begin{tabular}{@{}l|r@{}}
\toprule
     \multicolumn{2}{l}{ROVIOLI frame update} \\
  \hline
    ROVIO update & 22.7\,ms \\ 
    Feature tracking & 20.6\,ms \\ 
    Localization & 20.4\,ms \\ 
    Map building & 3.2\,ms \\ 
\hline
    \textbf{Total} & \textbf{44.2\,ms} \\
\bottomrule
\end{tabular}
\end{subtable}
\end{table}

\subsection{Multi-session mapping}
\label{cla_mapping}

In many mapping applications, it is not possible to cover the entire environment within a single mapping session.
Apart from that, it might be desirable to capture the environment in as many differing visual appearance conditions as possible~\cite{buerkiIROS2016}.
Therefore, maplab offers tools to co-register maps from multiple sessions together and jointly refine them to obtain a single, consistent map.

Hence this use case demonstrates the process of creating a map of a university building from 4 individual trajectories.
Each trajectory passes through the ground floor, staircases and one other floor of a building.
Combined, they cover over 1,000 meters and contain about 463,000 landmarks.
On such large maps, many of the common operations such as optimization or loop closure quickly become intractable without a careful selection of the data.
For this reason, we employ a keyframing scheme using heuristics based on vertex distance, orientation, and landmark covisibility.
The loop closure algorithm of maplab correctly identifies the geometric transformations between all missions and the non-linear optimization refines the geometry.
The result is a compact, geometrically-consistent localization map of 8.2~MB ready to be used by ROVIOLI for localization within the entire building as shown in \reffig{fig:teaser}.

%

This use case can be reproduced using the following commands in the maplab console:
\begin{lstlisting}[language=bash,commentstyle={\color{blue}},keywordstyle={\color{black}}]
# Load multiple single session maps from ROVIOLI.
load_merge_all_maps --maps_folder YOUR_MAPS_FOLDER
# Keyframing and initial optimization.
kfh
optvi
# Set one mission as base, anchor the others.
set_mission_baseframe_to_known
anchor_all_missions
# Pose-graph relaxation, loop-closure, optimization.
relax
lc
optvi
visualize
\end{lstlisting}

\subsection{Map maintenance}
Large feature-based models, potentially built in multiple sessions, easily comprise thousands of landmarks and reach considerable storage size.
However, it is not really necessary to keep all of the landmarks to guarantee good localization quality with ROVIOLI.
Maplab offers a map summarization functionality based on~\cite{dymczyk2015keep} that uses an integer-based optimization to perform the landmark selection.
The algorithm attempts to remove the least commonly seen landmarks but at the same time maintain a balanced coverage of the environment.
Maplab also includes a keyframing algorithm to remove redundant vertices and only keep the ones necessary for an efficient and accurate state estimation.
By removing the vertices we also eliminate many vertex-landmark associations that contain descriptors of considerable size.
Both summarization and keyframing permit to significantly reduce the model size without a large loss in pose estimation quality.

The map maintenance is demonstrated on a database map built from 4 mapping sessions recorded on the ground floor of the building introduced in~\refsec{cla_mapping}.
Each mapping session covers about 90~meters and contains about 20,000 landmarks, out of which about 5,000 are considered reliable.
A 5th dataset is used as a query -- we try to localize each vertex against the database, built from the 4 datasets, and verify if the position error is smaller than 50~cm.
We compare the recall of localization maps that were pre-processed in different ways, either summarized, keyframed or both.

\begin{figure}
    \centering
    \includegraphics[width=0.49\textwidth]{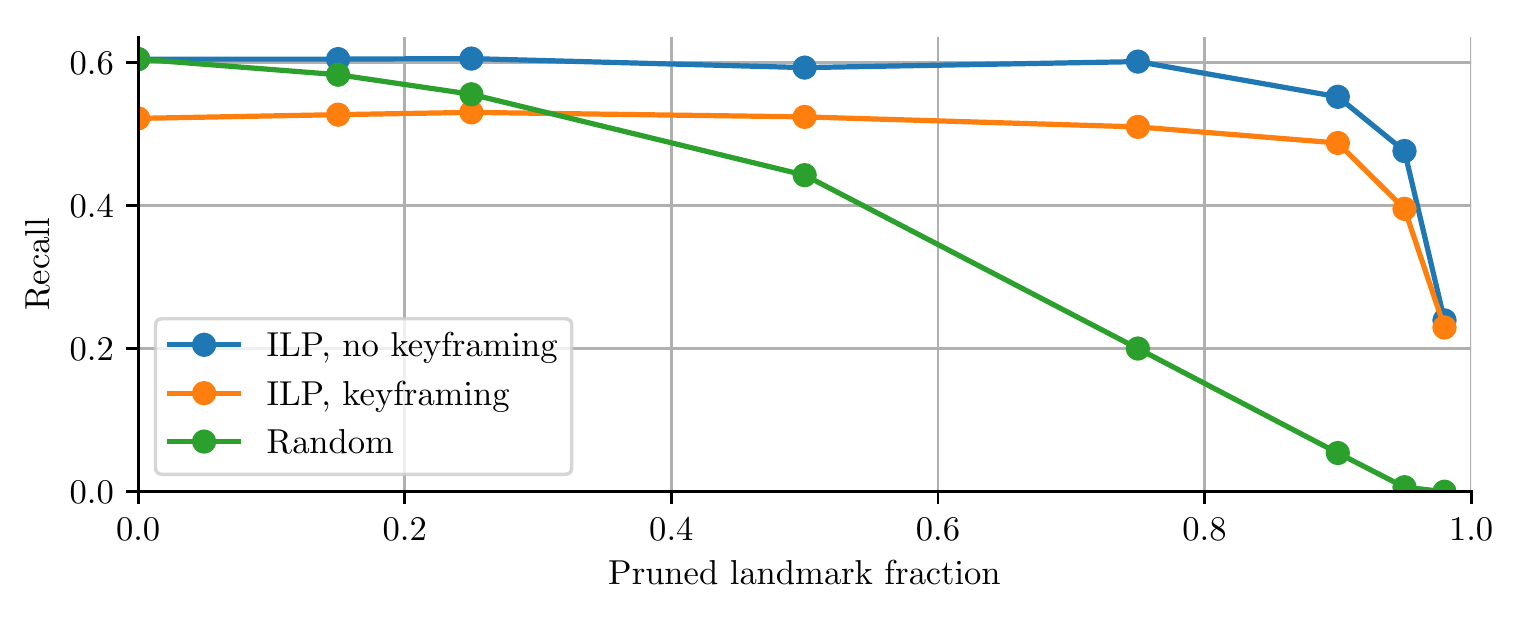}
    \scriptsize
    \begin{tabular}{@{}l|cccccc@{}}
        \toprule
        pruning fraction & 0 & 0.5 & 0.75 & 0.9 & 0.95 & 0.98 \\
        \hline
        \# landmarks &
            \begin{tabular}{@{}c@{}}18,316\\(16,088)\end{tabular} &
            \begin{tabular}{@{}c@{}}8,824\\(9,148)\end{tabular} &
            \begin{tabular}{@{}c@{}}4,259\\(4,570)\end{tabular} &
            \begin{tabular}{@{}c@{}}1,818\\(1,822)\end{tabular} &
            \begin{tabular}{@{}c@{}}899\\(906)\end{tabular} &
            \begin{tabular}{@{}c@{}}349\\(358)\end{tabular}\\
        \hline
        map size {[}MB{]} & 
        \begin{tabular}{@{}c@{}}34.559\\(3.740)\end{tabular} &
        \begin{tabular}{@{}c@{}}29.217\\(3.209)\end{tabular} &
        \begin{tabular}{@{}c@{}}24.028\\(2.602)\end{tabular} &
        \begin{tabular}{@{}c@{}}17.619\\(1.837)\end{tabular} &
        \begin{tabular}{@{}c@{}}12.203\\(1.214)\end{tabular} &
        \begin{tabular}{@{}c@{}}6.707\\(0.712)\end{tabular} \\
        \bottomrule
    \end{tabular}
    
    \caption{
    The localization performance and map size after ILP landmark summarization and keyframing+summarization (in brackets).
    Keyframing removes vertices including vertex-landmark associations, effectively making the map smaller. 
    The original map had $6,258$ vertices whereas the keyframed map contains $760$.
    Keyframing consistently reduces the recall by a few percent while summarization only affects the quality when the pruned landmark fraction exceeds 85\%.
    For comparison, we provide a recall curve for a random selection of landmarks to be removed.
    }
    \label{fig:summarization}
\end{figure}

\reffig{fig:summarization} presents the influence of landmark summarization and keyframing on the localization map size and demonstrates how those approaches affect the localization.
The results confirm that keyframing significantly reduces the localization map size with a rather marginal loss of localization quality.
Similarly, summarization can reduce the total amount of landmarks by 90\% without grave consequences.
When these methods are combined we can reduce the map size 13 times and keep the recall level at 51\%, compared to 60\% for the full map.

\begin{lstlisting}[language=bash,commentstyle={\color{blue}},keywordstyle={\color{black}}]
# Keyframe the map and sparsify landmarks to 10,000.
kfh
landmark_sparsify --num_landmarks_to_keep=10000
\end{lstlisting}

\subsection{Large-scale mapping}
\label{large_scale_mapping}

\begin{figure}
\centering
\includegraphics[width=1.0\linewidth]{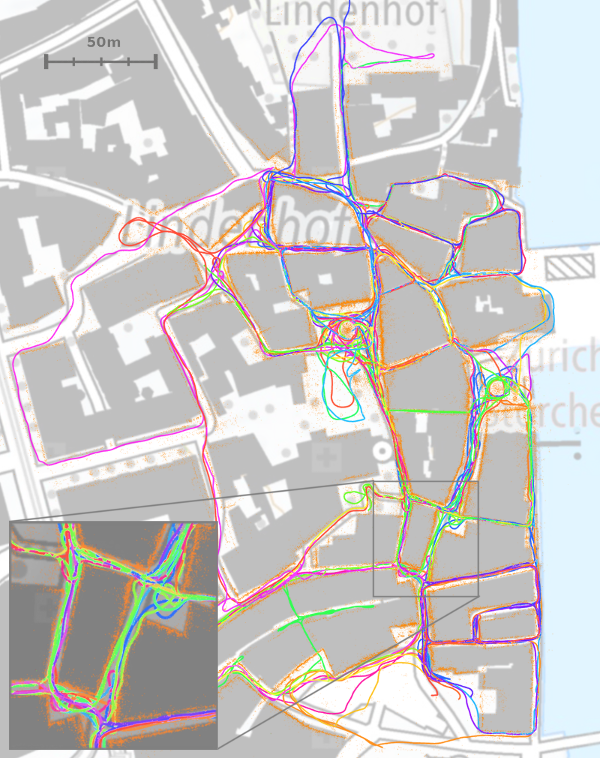}
\caption{
Large-scale, multi-session VI-map of Zurich's old town.
Built from the raw visual-inertial data recorded in $45$ sessions using Google Tango tablets on two different days (sunny and cloudy).
The total duration of the recordings is $231$\,min.
The final map contains trajectories with a total length of $16.48$\,km, $435$k landmarks with $7.3$M observations and has a size of $480$\,MB.
The map is available on the maplab wiki page for download.
}
\label{fig:old_town_topdown}
\end{figure}

In this use-case we would like to demonstrate the large-scale mapping capabilities of maplab and the applicability to a sensor other than the VI-sensor~\cite{nikolic2014synchronized}.
To that end we used the publicly available Google Tango tablets, and recorded a large-scale, multi-session map of the old town of Zurich.
We exported the raw visual-inertial data and processed it with ROVIOLI to obtain the initial open loop maps.
We then loaded these maps into the maplab console for alignment and optimization and applied the same tools as described in \refsec{cla_mapping}.
The bundle adjustment and pose-graph relaxation was performed on a desktop computer with $32$ GB RAM overnight.
An orthographic projection of the optimized VI-map onto the map of Zurich, as well as further details about the map can be found in \reffig{fig:old_town_topdown}.
The figure shows that the resulting map is consistent with the building and streets across most of the map with some minor inconsistencies in areas of low coverage.



\subsection{Dense reconstruction}
\label{dense_reconstruction}

Many applications in robotics, such as path planning, inspection and object detection require a more dense 3d representation of the environment.
Maplab offers several dense reconstruction tools, which use the optimized \textit{vertex} poses of the sparse map to compute dense depth information based on camera images attached to the VI-map.
\begin{figure*}
    \frame{\includegraphics[width=0.33\linewidth]{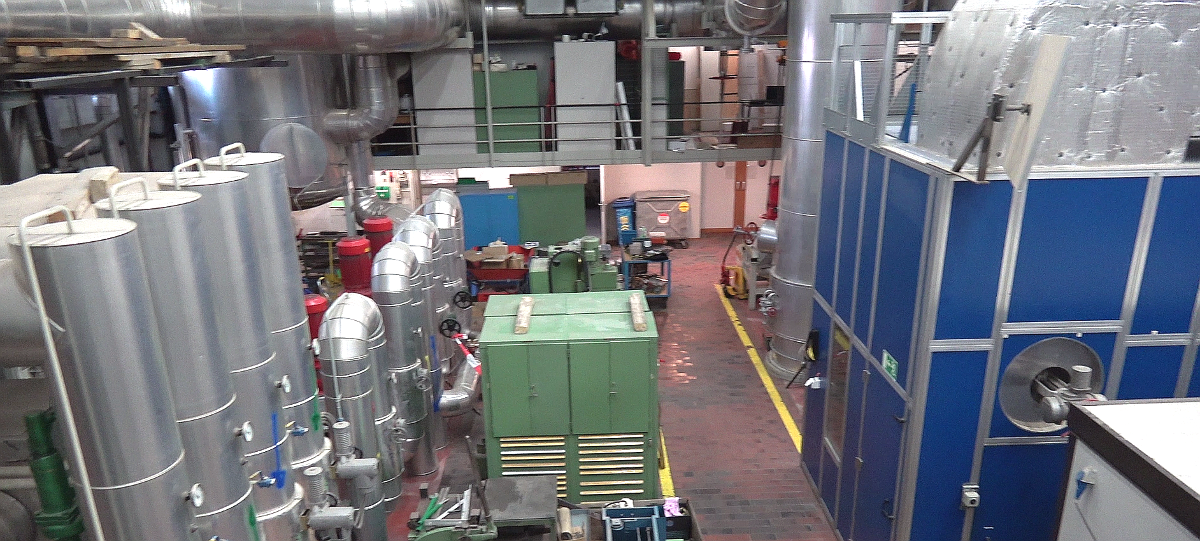}}
    \hspace*{-0.01\linewidth}
    \frame{\includegraphics[width=0.33\linewidth]{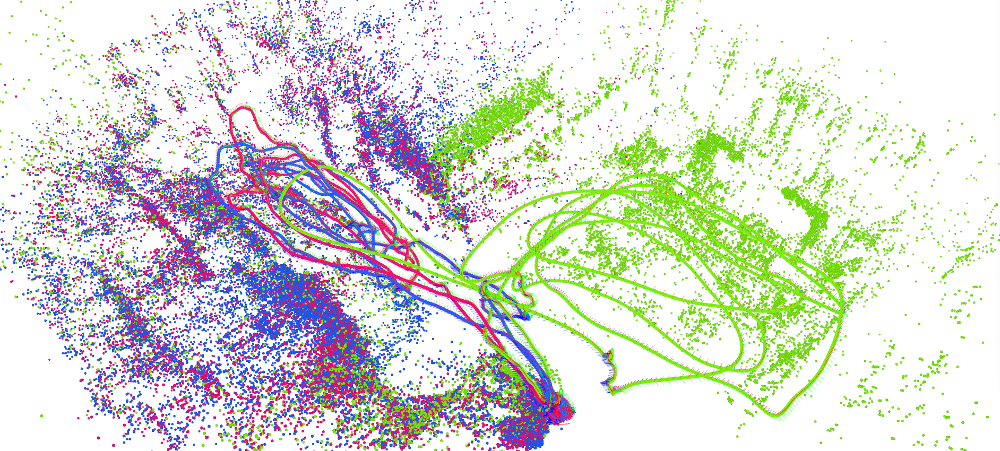}}
    \hspace*{-0.01\linewidth}
    \frame{\includegraphics[width=0.33\linewidth]{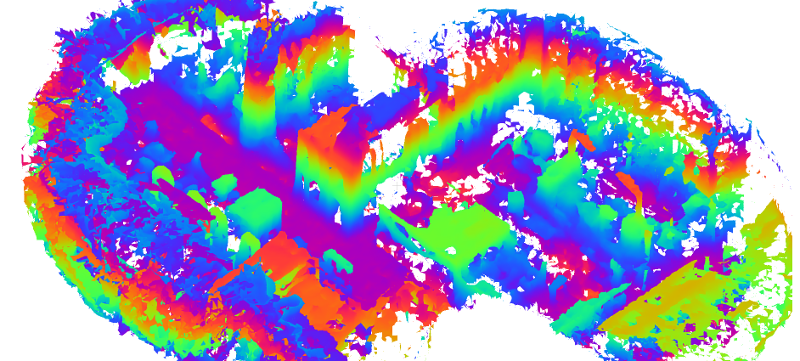}}

    \vspace*{0.01\linewidth}
    
    \frame{\includegraphics[width=0.33\linewidth]{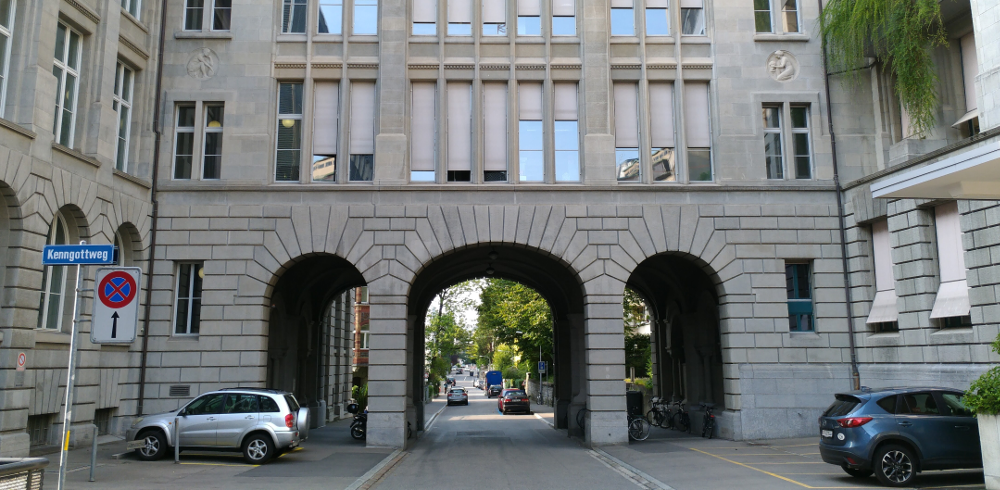}}
    \hspace*{-0.01\linewidth}
    \frame{\includegraphics[width=0.33\linewidth]{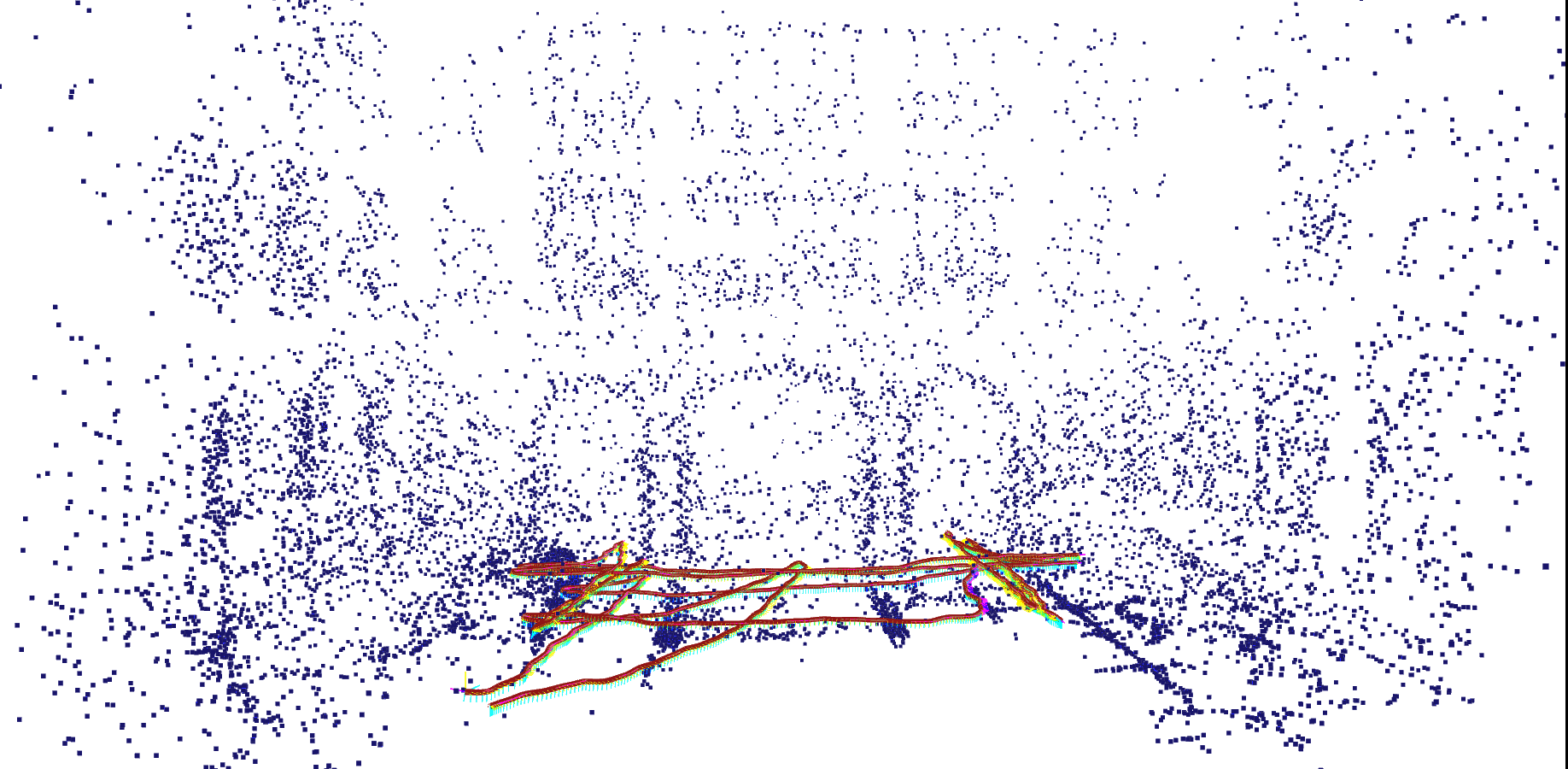}}
    \hspace*{-0.01\linewidth}
    \frame{\includegraphics[width=0.33\linewidth]{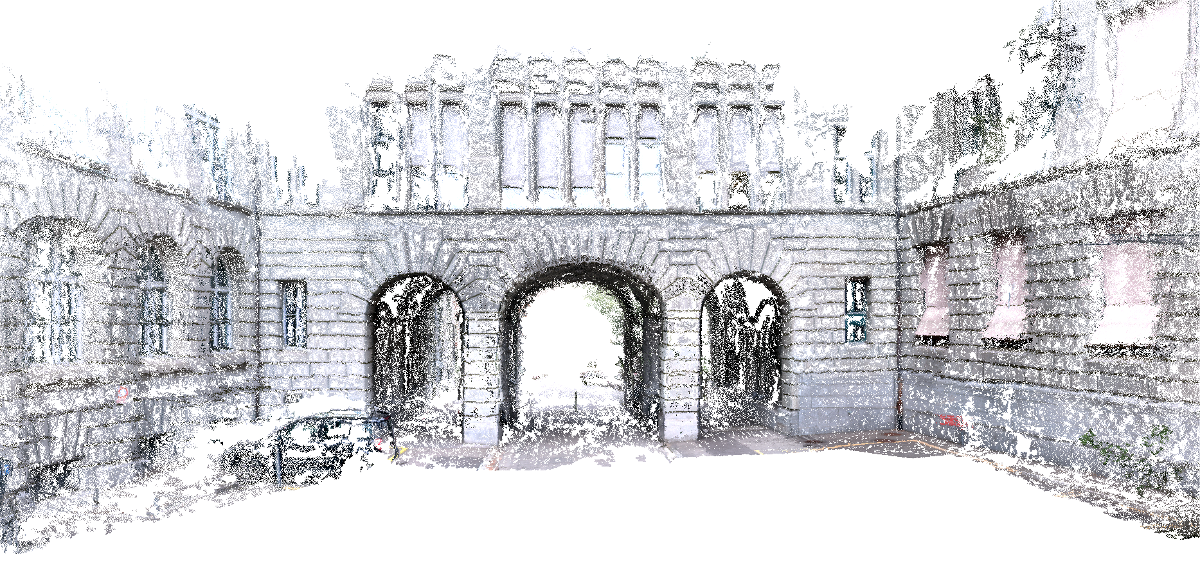}}
    \caption{Two different dense reconstruction tools are available in maplab.
    \textit{Top:} stereo dense reconstruction is used to compute depth maps based on grayscale images and optimized camera poses. They are then fused in voxblox~\cite{oleynikova2016voxblox} to create a surface mesh. 3 EuRoC datasets~\cite{burri2015real} (MH1-3) are combined to create an aligned and optimized VI-map.
    \textit{Bottom:} CMVS/PMVS2~\cite{Furu:2010} reconstruction results based on a single recording session using a multi-camera system with a RGB camera.}
    \label{fig:dense_reconstruction}
\end{figure*}
\subsubsection{Stereo dense reconstruction}
In order to compute depth maps from multi-camera systems, this tool first identifies stereo cameras that are suitable for planar rectification.
It then utilizes a (semi-global) block matcher to compute depth maps for every stereo pair along the trajectory.
The resulting depth maps (or point clouds) are attached to the VI-map and stored in the resource system.
The following commands assume that the maps are already aligned, loop-closed and optimized as described in~\refsec{cla_mapping}.
\begin{lstlisting}[language=bash,commentstyle={\color{blue}},keywordstyle={\color{black}}]
stereo_dense_reconstruction
\end{lstlisting}
\subsubsection{TSDF-based depth fusion}
Once the VI-map contains depth information, e.g. obtained using the above described commands or an RGB-D sensor, the globally consistent camera poses of the VI-map can be utilized to create an equally consistent global 3d reconstruction.
To that end, maplab employs voxblox~\cite{oleynikova2016voxblox}, a volumetric mapping library, for TSDF-based depth fusion and surface reconstruction.
The following commands will insert depth maps or point cloud data into a voxblox grid and store a surface mesh to the file system.
The top row of~\reffig{fig:dense_reconstruction} shows the reconstruction results of 3 combined EuRoC machine hall datasets~\cite{burri2016euroc}.
\begin{lstlisting}[language=bash,commentstyle={\color{blue}},keywordstyle={\color{black}}]
create_tsdf_from_depth_resource 
    --dense_tsdf_voxel_size_m 0.10
    --dense_tsdf_truncation_distance_m 0.30
export_tsdf
    --dense_result_mesh_output_file YOUR_FILE
\end{lstlisting}
\subsubsection{Export to CMVS/PMVS2}
For more accurate dense reconstructions maplab offers an export command to convert the sparse VI-map and images to the input data format for the open-source multi-view-stereo pipeline, \mbox{CMVS/PMVS2~\cite{Furu:2010}}.
Even though the export of grayscale images is supported, the best results are obtained using RGB images.
The VI-map and the resulting 3D reconstruction can be seen in the bottom row of~\reffig{fig:dense_reconstruction}.
\begin{lstlisting}[language=bash,commentstyle={\color{blue}},keywordstyle={\color{black}}]
export_for_pmvs
    --pmvs_reconstruction_folder EXPORT_FOLDER
\end{lstlisting}

\section{Using maplab for research}
All the algorithms and console commands required for the use-cases in Section~\ref{sec:maplab_usecases} are available in maplab and constitute most of the basic tools needed in visual-inertial mapping and localization.
Furthermore, a rich set of helper functions, queries, and manipulation tools are provided to ease rapid prototyping of new algorithms.
The plugin architecture of the console allows for an easy integration of new algorithms into the system.
Examples demonstrating how to extend the framework are provided in the project's wiki pages.
We would like to invite the community to take advantage of this research-friendly design.

\section{Conclusions}
%
%
This work presents maplab, an open framework for visual-inertial mapping and localization with the goal of making research in this field more efficient by providing a collection of basic algorithms and letting researchers focus on actual tasks.
All components in maplab are written in a flexible and extensible way such that novel algorithms that rely on visual-inertial state estimates or localization can be integrated and tested easily.
For this reason, the framework provides an implementation of the most important tools required in mapping and localization related research such as visual-inertial optimization, a loop-closure/localization backend, multi-session map merging, pose-graph relaxation and extensive introspection and visualization tools.
All these algorithms are made accessible from a console-based user interface where they can be applied to single or multi-session maps.
Such a workflow has proven to be very efficient when prototyping new algorithms or tuning parameters.

Secondly, the framework contains an online visual-inertial mapping and localization front-end, named ROVIOLI.
It can build new maps from raw visual and inertial sensor data and additionally track a global (drift-free) pose in real-time if a localization map is provided.
Previous work made use of this capability on different robotic platforms and demonstrated its ability of accurately tracking a global pose for a multitude of applications, including navigation and trajectory following.

\section{Acknowledgement}

We would like to acknowledge the many other contributors of maplab, most importantly: 
Titus Cieslewski, Mathias Bürki, Timo Hinzmann, Mathias Gehrig and Nicolas Degen.
Furthermore we would also like to thank Michael Blösch the author of ROVIO.
The research leading to these results has received funding from Google Tango and the EU H2020 project UP-Drive under grant no. 688652.

\bibliographystyle{ieeetr} 
\small
\bibliography{robotvision}


\addtolength{\textheight}{-1cm}
\end{document}